\documentclass{article}
\usepackage{graphicx} 
\usepackage[utf8]{inputenc}
\usepackage[english]{babel}
\usepackage{pdfpages}
\usepackage{amsfonts}
\usepackage{amsmath}
\usepackage{amssymb}
\usepackage{amsthm}
\usepackage{latexsym}
\usepackage{setspace}
\usepackage{mathtools}
\usepackage{stmaryrd}
\usepackage{hyperref}
\usepackage{fancyhdr}
\usepackage{csvsimple,booktabs} 
\usepackage{algpseudocode}
\usepackage{multirow}
\usepackage{subfigure}
\usepackage{float}
\usepackage{url}
\usepackage{adjustbox}
\usepackage{changepage}
\usepackage{multicol}
\usepackage{graphicx}
\usepackage{tikz}
\usepackage{lipsum}
\usepackage{subcaption}
\theoremstyle{plain}
\newtheorem{theorem}{Theorem}[section]

\theoremstyle{remark}

\theoremstyle{definition}
\newtheorem{definition}{Definition}[section]

\title{An approximation-based approach versus an AI one for the study of CT images of abdominal aorta aneurysms}
\author{
\hskip0.4cm {\bf L. Rinelli}$^a$, \hskip0.4cm {\bf A. Travaglini}$^{a,b,c}$, \hskip0.4cm {\bf N. Vescera}$^{a}$, \hskip0.4cm {\bf G. Vinti}$^{a,d}$   \\    \\
$^a$  Department of Mathematics and Computer Science \\
            University of Perugia \\      1, Via Vanvitelli, 06123 Perugia, Italy    \\   
{\small {\tt lucrezia.rinelli@gmail.com} \hskip0.5cm 
 {\tt arianna.travaglini@collaboratori.unipg.it}} \\ {\small {\tt nicolo.vescera@unipg.it} \hskip0.5cm   {\tt gianluca.vinti@unipg.it}}  \\ \\
$^b$  Department of Mathematics and Computer Science “U.Dini” - DIMAI \\ University of Florence \\      67/a, Viale Giovanni Battista Morgagni, 50134 Firenze, Italy
 \\  
{\small {\tt \tt arianna.travaglini@unifi.it}} 
\\ \\
$^c$  Department of Mathematics, Computer Science and Economics \\ University of Basilicata \\ 10, V.le
dell’Ateneo Lucano, 85100 Potenza, Italy
 \\  
{\small {\tt \tt arianna.travaglini@unibas.it}} 
\\ \\
$^d$ Department of Mathematics and Informatics \\ Lucian Blaga University of Sibiu \\ 5-7, Str. Dr. I. Ratiu, RO-550012 Sibiu, Romania
}
\date{}
\begin{document}

\maketitle
\begin{abstract}
    This study evaluates two approaches applied to computed tomography (CT) images of patients with abdominal aortic aneurysm: one deterministic, based on tools of Approximation Theory, and one based on Artificial Intelligence. Both aim to segment the basal CT images to extract the patent area of the aortic vessel, in order to propose an alternative to nephrotoxic contrast agents for diagnosing this pathology. While the deterministic approach employs sampling Kantorovich  operators and the theory behind, leveraging the reconstruction and enhancement capabilities of these operators applied to images, the artificial intelligence-based approach lays on a U-net neural network. The results obtained from testing the two methods have been compared numerically and visually to assess their performances, demonstrating that both models yield accurate results.
\end{abstract}
\section{Introduction}
The aim of this work is to propose a comparison between a deterministic approach (DA), based on Approximation Theory, and an Artificial Intelligence-based approach (AI), applied to the study of CT images of patients affected by abdominal aortic aneurysm (AAA). \\This pathology, widespread, consists of an abnormal dilation of the aortic vessel, diagnosed when the vessel diameter exceeds $50\%$ of normal conditions. Multiple factors contribute to the development of an aortic aneurysm, which in most cases is an asymptomatic condition until it progresses to the most severe complication, which is the rupture. This condition carries a high mortality rate, with approximately $79\%$ of patients dying before reaching the hospital (see e.g. \cite{4}).  The gold standard procedure for studying AAA is computed tomography (CT). The goal of the diagnosis is to identify the patency of the aortic vessel in the aneurysmal area, because this condition typically tends to reduce it. However, since blood is nearly radio-transparent, it cannot be distinguished in a basal CT image. Therefore, contrast-enhancing substances, known as contrast agents, are used. The diagnostic procedure involves an initial basal scan, followed by the administration of the contrast agent to the patient in appropriate doses, and then a second scan is executed, providing the contrast medium (CM) sequence, which clearly distinguishes the portion of the vessel where blood flows. The basal image is useful for identifying internal structures of the vessel such as calcium plaques. The primary concern regarding contrast agents is their high nephrotoxicity, rendering them unsuitable for administration to patients with renal issues or specific allergies (see e.g. \cite{fukushima}). Consequently, we work with Image Processing methods aimed at simulating the effects of contrast agents on CT images without actually using them, considering only basal images. The objective is to enable doctors to identify the patent lumen of the vessel in case of AAA.\\ We will analyze two approaches: the first is based on sampling theory.\\ Specifically, here the central role is played by the sampling Kantorovich (SK) operators, introduced in the one-dimensional setting in \cite{kant_in_orl_space}, and later extended to the multidimensional case in \cite{bumi}, where they are defined as follows:
\begin{equation}
\label{def_sk_intro}
        (S_w^\chi f)(\underline{x})= \sum_{\underline{k}\in\mathbb{Z}^n}\chi(w\underline{x}-t_{\underline{k}})\cdot \bigg[\frac{w^n}{A_{\underline{k}}}\cdot \int_{R_{\underline{k}}^w}f(\underline{u})d\underline{u}\bigg] \ \ \ (\underline{x}\in\mathbb{R}^n, w>0).
    \end{equation}
    In (\ref{def_sk_intro}) $\chi$ is the so-called kernel of the operator and $f:\mathbb{R}^n\to\mathbb{R}$ is a locally integrable function such that the above series converges for each $\underline{x}\in\mathbb{R}^n$. Moreover, $(t_{\underline{k}})_{\underline{k}\in\mathbb{Z}^{n}}$ is a suitable sequence of values, $R_{\underline{k}}^{w}$ is the plurirectangle representing a neighborhood of the current sample value and $\frac{A_{\underline{k}}}{w^n}$ is the Lebesgue measure of $R_{\underline{k}}^{w}$. Finally, $w$ is the sampling rate. All these notations will be explained later. 
The family of SK operators distinguishes itself from previous family of generalized sampling operators (see e.g. \cite{butzer}) by replacing the sampled values of the signal $f$ with the integral mean on $R_{\underline{k}}^{w}$. Consequently, the operator inherits the regularization properties of the integral mean, which also helps in the reduction of time-jitter error, typical in sampling theory, that occurs when it is not possible to sample the signal at a precise value $\frac{t_{\underline{k}}}{w}$.\\
These operators, due to their nature, act simultaneously as a low pass filter, reducing noise and as rescaling algorithm with better performances respect to other algorithms used with this aim (see e.g. \cite{comparison, alz}). These operators were integrated into a segmentation algorithm \cite{segmentation} that extracts the patent zone of the aortic vessel starting from the basal image. This algorithm was implemented within a portal called ImageLab \cite{travaglini}, that was used for tests. Specifically, compared to the version presented in \cite{travaglini}, the algorithm includes remarkable improvements that enhance its performance, such as additional Digital Image Processing operations and the removal of calcium plaques.
\\The second approach lays on artificial intelligence. We trained a U-Net neural network that in the testing phase takes as input the basal mode image and provides as output the prediction of the CM one. \\ The obtained results show a good efficiency of the two models used for the extraction of the patent lumen of the aortic vessel. The tests carried out have been compared using similarity indices, showing a slight advantage in the performance of the network with respect to that of the deterministic algorithm.\\ We would like to point out that the CT images used in this work, which have been kindly granted by the Division of Diagnostic Imaging of the Department of Medicine and Surgery of the University of Perugia, are covered by an informed consent, signed by the patients involved, and are completely anonymized.\\ \\ In Section 2, we review fundamental concepts and notation pertaining to SK operators. Section 3 introduces the deterministic approach (DA), highlighting the theory behind the mathematical model, the segmentation algorithm and similarity indices used for evaluation. Additionally, we outline the dataset employed and detail the obtained results. Section 4 is devoted to the AI approach: we describe the model of the used U-Net, the datasets and the obtained results. In Section 5, we conduct a comparative analysis of the results obtained from the two approaches to determine their respective efficiency.
Finally, in Section 6 we outline the conclusions.

\section{Preliminaries and approximation results}
In order to apply the multivariate sampling Kantorovich operators to images, it is necessary to recall some preliminary notions.\\
Let $\Pi^n=(t_{\underline{k}})_{\underline{k}\in\mathbb{Z}^n}$ be a sequence defined by $t_{\underline{k}}=(t_{k_1}, ..., t_{k_n})$, where $(t_{k_i})_{k_{i}\in\mathbb{Z}}, \ \ i=1,...,n$ is a sequence of real numbers for which $-\infty<t_{k_i}<t_{k_{i+1}}<+\infty$, $\lim_{k_i\to\pm\infty}t_{k_i}=\pm\infty$ for every $i=1,...,n$ and such that there exist $\Delta, \ \delta>0$ for which $\delta\leq\Delta_{k_i}=t_{k_{i}+1}-t_{k_i}\leq \Delta$, for every $i=1,...,n$.\\
Then, a function $\chi:\mathbb{R}^n \to \mathbb{R}$ is called a \textit{kernel} (for the sampling Kantorovich operator) if it satisfies the following properties:
\begin{itemize}
    \item (K1) $\chi \in L^1(\mathbb{R}^n)$ and is bounded in a neighbourhood of $\underline{0}\in\mathbb{R}^n$;
    \item (K2) For every $\underline{u}\in\mathbb{R}^n$, $$\sum_{\underline{k}\in\mathbb{Z}^n}\chi(\underline{u}-t_{\underline{k}})=1;$$
    \item (K3) For some $\beta>0$, $$m_{\beta,\Pi^n}(\chi)=\sup_{\underline{u}\in\mathbb{R}^n}\sum_{\underline{k}\in\mathbb{Z}^n}|\chi(\underline{u}-t_{\underline{k}})|\cdot ||\underline{u}-t_{\underline{k}}||_{2}^{\beta}< +\infty,$$
    where $m_{\beta,\Pi^n}(\chi)$ denotes the absolute discrete moment of order $\beta$.
\end{itemize}
We define the set 
\begin{equation}
R_{\underline{k}}^w=\bigg[\frac{t_{k_1}}{w},\frac{t_{{k_1}+1}}{w} \bigg]\times \bigg[\frac{t_{k_2}}{w}, \frac{t_{{k_2}+1}}{w}\bigg]\times ... \times \bigg[\frac{t_{k_n}}{w}, \frac{t_{{k_n}+1}}{w}\bigg] \ \ \ \ (w>0).
\end{equation}
For every $\underline{k}\in\mathbb{Z}^n$, and $w>0$, $R_{\underline{k}}^w$ is a subset of $\mathbb{R}^n$. \\If we denote by $A_{\underline{k}}:= \Delta_{k_1}\cdot ... \cdot \Delta_{k_n}$, then the Lebesgue measure of $R_{\underline{k}}^w$ is given by $\frac{A_{\underline{k}}}{w^n}$ and $$\frac{\delta^n}{w^n}\leq \frac{A_{\underline{k}}}{w^n}\leq \frac{\Delta^n}{w^n}.$$
\begin{definition}
    Let $\chi$ be a kernel. The family of operators $(S_w^\chi)_{w>0}$ defined by
    \begin{equation}
    \label{def_sk}
        (S_w^\chi f)(\underline{x})= \sum_{\underline{k}\in\mathbb{Z}^n}\chi(w\underline{x}-t_{\underline{k}})\cdot \bigg[\frac{w^n}{A_{\underline{k}}}\cdot \int_{R_{\underline{k}}^w}f(\underline{u})d\underline{u}\bigg] \ \ \ \underline{x}\in\mathbb{R}^n,
    \end{equation}
    where $f:\mathbb{R}^n\to\mathbb{R}$ is a locally integrable function such that the series converges for each $\underline{x}\in\mathbb{R}^n$, is called the \textit{multivariate sampling Kantorovich operator}, briefly SK-operator. 
\end{definition}
In what follows we report some approximation results, proved in \cite{bumi}, that plays a crucial role in the application of the SK operators to digital image processing. Indeed, the implementation of these results allows the reconstruction and the enhancement of the images. \\In particular, when dealing with continuous signals, the following convergence result holds.
\begin{theorem}
    \label{teorema4.1}
    Let $f\in C^0(\mathbb{R}^n)$. Then, for every $\underline{x}\in\mathbb{R}^n$, $$\lim_{w\to + \infty}(S_{w}^{\chi}f)(\underline{x})=f(\underline{x}).$$
    In particular, if $f\in C(\mathbb{R}^n)$, then $$\lim_{w\to +\infty}||S_{w}^{\chi}f-f||_{\infty}=0,$$
where $C(\mathbb{R}^{n})$ (resp. $C^{0}(\mathbb{R}^n)$) denotes the space of uniformly continuous and bounded (resp. continuous and bounded) functions on $\mathbb{R}^{n}$.
\end{theorem}
Another noteworthy outcome shows that it is possible to reconstruct by the SK operators even not-necessarily continuous signals. Indeed, the following $L^{p}$-convergence theorem holds. 
\begin{theorem}
\label{convergenza_sk}
    For every $f\in L^{p}(\mathbb{R}^n)$, $1\leq p< +\infty$, we have $$\lim_{w\to +\infty}||S_{w}^{\chi}f-f||_{p}=0.$$
    Moreover, the following inequality holds $$||S_{w}^{\chi}||_{p}\leq \delta^{-n/p}(m_{0, \Pi^n}(\chi))^{(p-1)/p}||\chi||_{1}^{1/p}||f||_{p} \ \ \ (f\in L^{p}(\mathbb{R}^n)).$$
\end{theorem}
For further results about SK operators, see e.g. \cite{bardaro_vinti, bumi, approx_applied, mantellini, zampogni2018, sambucini, cost_vinti_2019, acar, cma, BS2, regul_sat, zampogni2022, BS1}.\\ \\
A fundamental role in the SK operators and their implementation is played by the kernel, i.e., the function that satisfies conditions (K1)-(K3). In literature, there are several examples of such functions, below described.\\
In the segmentation process conducted in this work, it has been chosen the \textit{Jackson-type kernels} for the SK operators. These are a class of product-type (tensor product) multivariate kernels (see e.g. \cite{butzer_fourier}), defined as:
\begin{equation}
    \label{jackson_multi}
    \mathcal{J}_{k}^{n}(\underline{x})= \prod_{i=1}^{n}J_{k}(x_i),\ \ \ \underline{x}\in\mathbb{R}^n,
    \end{equation}
in which $$J_{k}(x)=c_{k}sinc^{2k}\bigg(\frac{x}{2k\pi \alpha}\bigg)$$ are the one-dimensional Jackson-type kernels, with $x\in\mathbb{R}, \ k\in\mathbb{N}, \ \alpha\geq 1$, where
$$sinc(x):=\begin{cases} \frac{sin(\pi x)}{\pi x} \ \ \ x\in\mathbb{R} \hspace{-0.05cm}\smallsetminus\hspace{-0.05cm}  \{0\} \\ 1 \hspace{1.2cm} x=0,\end{cases}$$ and $c_k$ is a non-zero normalization coefficient given by $$c_{k}:= \bigg[\int_{\mathbb{R}}sinc^{2k}\bigg(\frac{u}{2k\pi\alpha}\bigg)du\bigg]^{-1}.$$ \\
Another class of product-type kernels is the one of \textit{de la Vallée Poussin's kernels}, defined as $$\theta_{n}(\underline{x}):=\frac{4}{(\sqrt{2\pi})^{n}}\prod_{i=1}^{n}\frac{sin(x_{i}/2)sin(3x_{i}/2)}{x_{i}^2}, \ \ \ x\in\mathbb{R}^n,$$ where $$\theta(x)=\frac{2}{\pi}\frac{sin(x/2)sin(3x/2)}{x^2}, \ \ \ x\in\mathbb{R}$$ is its one-dimensional version.\\
Other multivariate kernels that are not of product type, are the radial-type ones. For example, we recall the \textit{Wendland kernels} (see e.g.\cite{wend_95, wend_04}), which are compactly supported radial basis functions. These are defined by means of the function 
\begin{equation}
\phi_{m,0}(r):=(1-r)_{+}^{m}:=
\begin{cases}
    (1-r)^m, \hspace{0.5cm} if \ \ r\leq1, \\
    0, \hspace{0.5cm} if \ \ r>1,
\end{cases}
\text{with} \hspace{0.3cm} r=||\underline{x}||_{2} \hspace{1.0cm} (\underline{x}\in\mathbb{R}^n).
\end{equation}
Then, the Wendland kernels are defined by the recursive formula $$\phi_{m, k+1}(r):=\int_{r}^{+\infty}t\phi_{m,k}(t)dt, \hspace{0.5cm} k=0,1,...,$$
with $r=||\underline{x}||_{2}, \ \underline{x}\in\mathbb{R}^n$.
For more details about kernels, see e.g. \cite{butzer_fourier, gen_sam_approx}.

\section{Deterministic approach (DA)}
In this section we analyze the deterministic approach of our work, based on an algorithm that lead to the extraction of the pervious area of the aorta artery on CT basal images. This algorithm is implemented in a portal named \textit{ImageLab}, located at the Department of Mathematics and Computer Science of the University of Perugia. 
\subsection{Description of the mathematical model}
\begin{figure}[H]
\centering
\includegraphics[width=2.5cm]{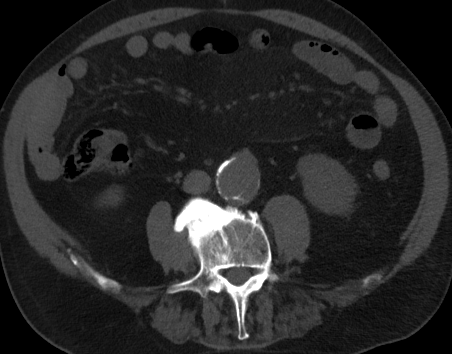}
\includegraphics[width=5cm]{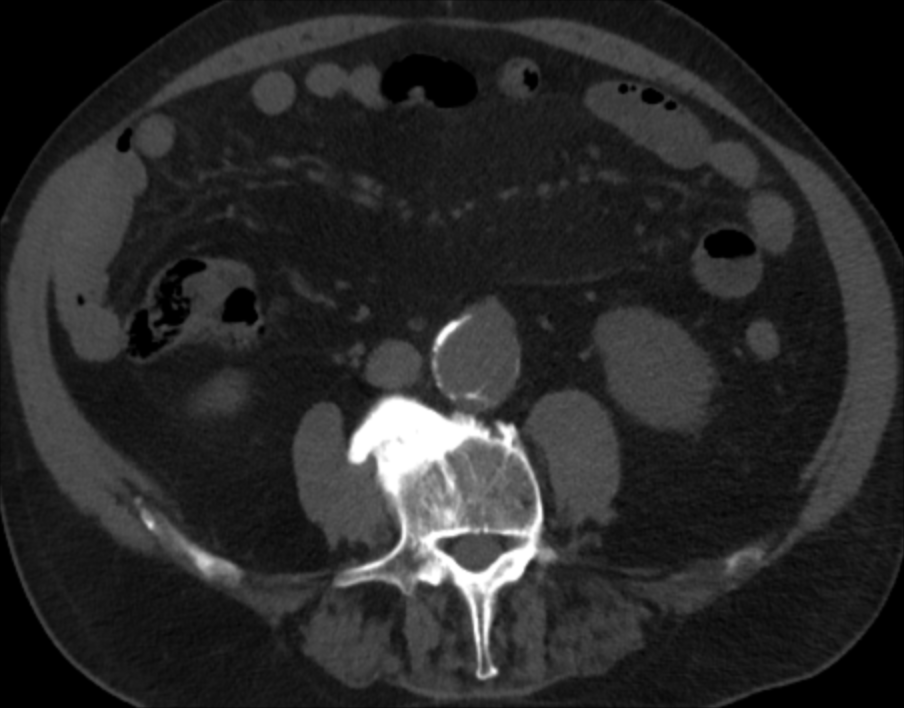}
\caption{\small{Reconstruction by SK operators, with the bidimensional Jackson type kernel $J^{2}_{12}$, $w=20$ and scaling factor $R=2$}. On the left we can see the original image, on the right the reconstructed one.}
    \label{ricostruzione_sk}
\end{figure}

\begin{figure}[H]
    \centering
    \includegraphics[width=13cm, keepaspectratio]{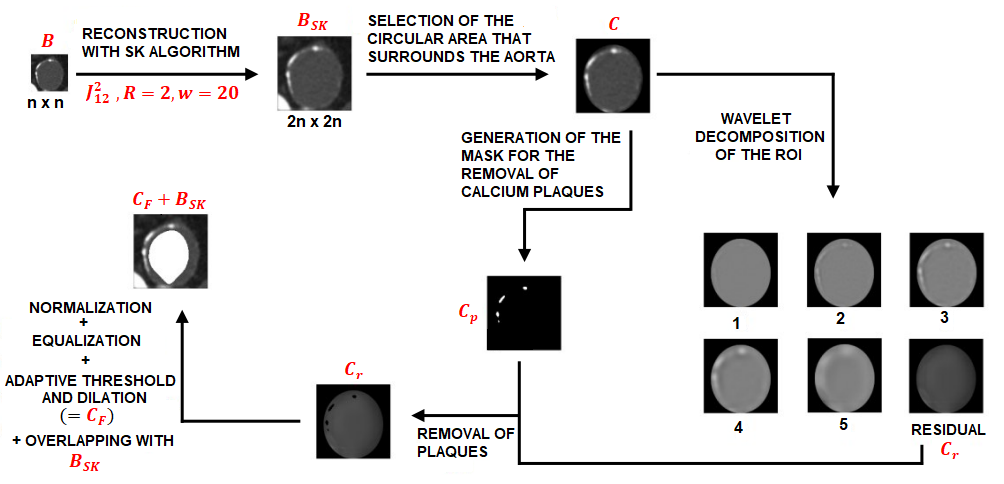}
    \caption{\small{Scheme of the segmentation algorithm.}}
    \label{schema_imageLab}
\end{figure}

We recall that a bi-dimensional gray-scale image is a matrix $A=(a_{ij})_{i,j=1,...,m}$ that can be represented by a step function $I \in L^p(\mathbb{R}^2)$, with $1\leq p<+\infty$.
\\In particular, 
\begin{equation}
\label{image_function}
I(x,y):= \sum_{i=1}^{m}\sum_{j=1}^{m}a_{ij}\cdot \mathbf{1}_{ij}(x,y) \ \ \ \ (x,y)\in\mathbb{R}^2
\end{equation}
where $$\mathbf{1}_{ij}(x,y)=\begin{cases}
    1, \ \ \ if \ \ (x,y)\in(i-1,i]\times (j-1,j], \\ 0, \ \ \ otherwise.
\end{cases}$$
With this background, we can use the function in (\ref{image_function}) to compute the family of bi-variate SK operators (with an appropriate kernel $\chi$), which will approximate $I$ pointwise at the continuity points and in $L^p$-sense (see Theorem \ref{teorema4.1} and \ref{convergenza_sk}). This procedure will reconstruct and enhance the original image $I$.
It is interesting to note that increasing the sampling rate $w$ enhances the accuracy of the reconstruction. Upon this procedure, an algorithm has been developed in \cite{segmentation} (see also \cite{comparison}), and an illustration of its application in reconstruction can be observed in Figure \ref{ricostruzione_sk}.
\vspace{-0.1cm}
The SK algorithm acts both as a low-pass filter, reducing possible disturbances caused by noise and as a rescaling algorithm, increasing the informative content of the image. Indeed, it has been experimentally observed to yield more accurate results than other algorithms, see e.g. \cite{comparison, alz, retiniche}.
Moreover the SK algorithm  is pivotal in the segmentation algorithm implemented in the portal (see \cite{travaglini}), that we resume below.\\
The goal is to extract the patent lumen of the vessel by segmenting the basal image. \\ As we can see in Figure \ref{schema_imageLab}, the algorithm takes as input the region of interest (ROI) image containing the aorta (image $B$), and obtained as described in the following subsection. The ROI is processed using the SK algorithm, with a rescaling factor of $R=2$, a bi-variate Jackson-type kernel $J_{12}^{2}$ and a sampling rate of $w=20$, obtaining the image $B_{SK}$. Then, the area that surrounds the aorta is selected, setting the pixels outside this selection to black, producing the image $C$. Subsequently, the image undergoes a wavelet decomposition (see e.g. \cite{dutilleux, gonzalez_wavelet}), dividing the image into five levels, each with its own frequency content. The purpose of this decomposition is to obtain the \textit{residual image} ($C_{r}$), which contains the lowest frequencies of the ROI. One of the improvements made to this version of the algorithm, compared to the one described in \cite{travaglini}, is represented by the removal of calcium plaques from both the basal and the CM images which produces remarkable outcomes on the extraction of the patent lumen of the vessel. In order to remove structures from the ROI that should not be included in the prediction of the pervious lumen, such as calcium plaques inside the vessel, a binary mask $C_{p}$ is generated in such a way that these structures are highlighted in white. The residual image is subtracted from this mask, followed by normalization and equalization (see e.g. \cite{gonzalez}) to enhance the contrast. Subsequently, an adaptive threshold is applied followed by a dilation operation, resulting in a binary image $C_{F}$ where pixels belonging to the extracted pervious area are represented in white. Finally, this binary image is superposed to the basal one reconstructed with the SK algorithm $(C_{F}+B_{SK})$, see Figure \ref{schema_imageLab}.\\
Moreover, we have the removal of calcium plaques also from the CM image, i.e. the target. This procedure involves rescaling the CM image to match the dimensions of the binary mask. It's worth highlighting that the latter image $(C_{p})$ was derived from the basal one, which was doubled in size during the reconstruction with the SK algorithm. Subsequently, this mask is subtracted from the CM image. 
\\For evaluating the algorithm's performance, a thresholding operation is conducted with a threshold value $\eta=127$, which corresponds to the mean value of the range $[0,255]$; from here, we obtain the binary image of the CM without calcium plaques, denoted as $CM_{b}$. \\ \\
To assess the accuracy of the algorithm, several similarity indices are computed.\\
The \textit{Dice coincidence index}  (DCI) (see, e.g., \cite{dice}) measures the similarity between two binary sets of points, and is defined as:
\begin{equation}
\label{dci}
    DCI:=\frac{2\cdot \#(C_{F}\land CM_{b})}{\#C_{F}+\#CM_{b}},
\end{equation}
where $\#C_{F}$ is the total number of white pixels in the binary image $C_{F}$ (see Figure \ref{schema_imageLab}), $\#CM_{b}$ is the number of white pixels in $CM_{b}$, and $\#(C_{F}\land CM_{b})$ is the number of white pixels in both the previous images.
\\With the same symbols, the \textit{Tanimoto index }(TI) (see, e.g.,\cite{duda}) measures the ratio between the correctly classified pixels and the total number of pixels in the reference and extracted area. It is defined as: 
\begin{equation}
    \label{ti}
    TI:=\frac{\#(C_{F}\land CM_{b})}{\#C_{F}+\#CM_{b}-\#(C_{F}\land CM_{b})}.
\end{equation}
The \textit{Misclassification error} evaluates the number of misclassified pixels between the extracted zone and the reference set:
\begin{equation}
    \label{em}
    E_{m}:=\frac{\#m}{\#C_{F}+\#CM_{b}-\#(C_{F}\land CM_{b})},
\end{equation}
where $\#m$ is the number of the wrongly classified pixels.
\\To evaluate if the algorithm tends to overestimate or underestimate, a useful index is the \textit{Bias} term, $B_{pn}$ (see, e.g. \cite{xie}):
\begin{equation}
    \label{bpn}
    B_{pn}:=\frac{\#fp-\#fn}{\#tp},
\end{equation}
where $fp$ is the set of false positive pixels, $fn$ is the set of false negative, and $tp$ is the set of true positive. If $B_{pn}>0$, it means that the algorithm tends to overestimate, otherwise it underestimates the pervious zone.
\\Lastly, we refer to another similarity index that will be used later in the procedure.\\The \textit{Structural Similarity index} (SSIM) is defined as: 
 \begin{equation}
            \label{SSIM}
            \begin{split}
            SSIM (X, Y)&= l(X,Y)\cdot c(X,Y) \cdot s(X,Y) \\ &= \frac{(2\mu_{X}\mu_{Y}+C_{1})(2\sigma_{XY}+C_{2})}{(\mu_{X}^{2}+\mu_{Y}^{2}+C_{1})(\sigma_{X}^{2}+\sigma_{Y}^{2}+C_{2})}
            \end{split}
        \end{equation}
where, given two images $X$ and $Y$, $\mu_{X}$ and $\mu_{Y}$ are the mean of the intensity values of the respective images, $\sigma_{X}^{2}, \sigma_{Y}^{2}$ are the variances of the images $X$ and $Y$, $\sigma_{XY}$ is the covariance between $X$ and $Y$, and $C_{1}, C_{2}$ are constants to prevent division by zero. It offers an objective assessment of the similarity between two images. SSIM integrates three key components of images, which are compared between the two images being evaluated: \textit{luminance} $l(X, Y)$, \textit{contrast} $c(X, Y)$, and \textit{structure} $s(X, Y)$. Here, the \textit{structure} component evaluates the correlation of local structures in images, considering the similarity of patterns, textures and details.
\\For more details about SSIM, see e.g. \cite{ssim}. 

\subsection{Description of the dataset}
\begin{figure}[H]
    \centering
    \includegraphics[width=10cm, keepaspectratio]{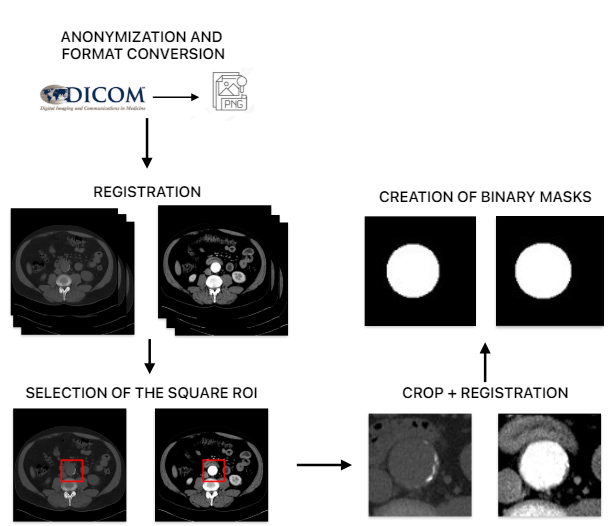}
    \caption{\small{Pre-processing steps of the dataset.}}
    \label{pre_elab}
\end{figure}

The procedure described in Fig. \ref{schema_imageLab} has been implemented in the ImageLab portal, which we used for testing. In the following, we describe the dataset and the necessary pre-elaboration steps before executing the processing in the portal. \\Native data are sequences of CT images in basal mode and with contrast medium (CM), acquired, in Hounsfield scale, in $12$ bit DICOM format. 
The images used have been provided by the Division of Diagnostic Imaging of the Department of Medicine and Surgery of the University of Perugia, belonging to six patients with AAA and to whom the contrast medium could be administered; furthermore, prior to undergoing the CT scan, they have signed an informed consent.  The thikness of the slices is about $1.0-2.0$mm, and the parameters of the windowing are $W=750 $ and $L=200$ for the basal sequence, $W=250$ and $L=75$ for the contrast medium one. 
As we can see from Figure \ref{pre_elab}, the sequences have been anonymized, by converting the images from DICOM to the lossless portable network graphics (.png) format, $512\times512$ pixel resolution, $8$ bit grey-levels. 
\\After that, from the entire stack of images is extracted the sequence containing the anatomical zone of interest, i.e., the part in which the aneurysm develops. 
\\Next, the sequences of images undergo to a \textit{registration algorithm} 
described by Philippe Thevenaz et al. in \cite{registration}.
\\Subsequently, images are cropped, in order to obtain a square image of dimensions $96\times 96$ (\textit{ROI}), which will include the vessel through the entire stack. These latter images are registered once again to improve the accuracy of the operation, which in some cases produces black edges on the registered image. However, these borders do not affect the testing procedure, as it focuses on the area representing the vessel. Finally, the sequence is saved in png format.

\begin{figure}[H]
    \centering
    \begin{tabular}{c c}
    \includegraphics[width=7cm]{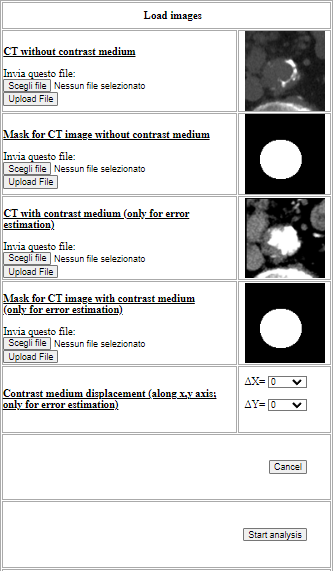} &
    \includegraphics[width=6.5cm]{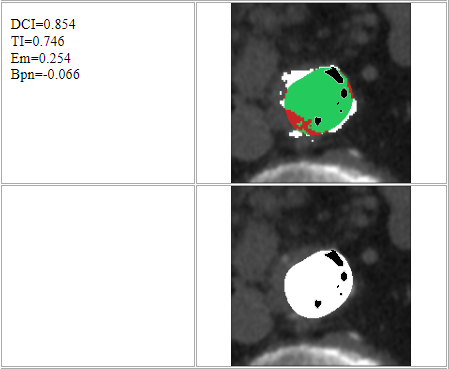}
    \end{tabular}
    \caption{\small{Interface and output of the portal ImageLab. We highlighted in black the calcium plaques.}}
    \label{interfaccia}
\end{figure}

\begin{figure}[H]
\hspace{-1.5cm}
    \begin{tabular}{c c  c c  c c}
      \includegraphics[width=2.0cm]{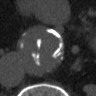} & \hspace{-0.2cm} \includegraphics[width=2.0cm]{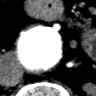} & \hspace{0.3cm} \includegraphics[width=2.0cm]{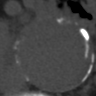} & 
      \hspace{-0.2cm}
      \includegraphics[width=2.0cm]{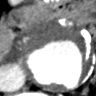} & 
      \hspace{0.3cm}
      \includegraphics[width=2.0cm]{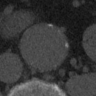} & 
      \hspace{-0.2cm}
      \includegraphics[width=2.0cm]{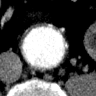}\\
      \multicolumn{2}{c}{P14: $45$ slices} & \multicolumn{2}{c}{P19: $49$ slices} & \multicolumn{2}{c}{P20: $76$ slices}\\ \\
      \includegraphics[width=2.0cm]{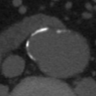} & 
      \hspace{-0.2cm}
      \includegraphics[width=2.0cm]{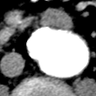} & 
      \hspace{0.3cm}
      \includegraphics[width=2.0cm]{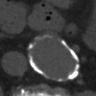} &
      \hspace{-0.2cm}
      \includegraphics[width=2.0cm]{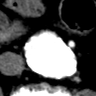} & 
      \hspace{0.3cm}
      \includegraphics[width=2.0cm]{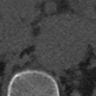} & 
      \hspace{-0.2cm}
      \includegraphics[width=2.0cm]{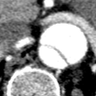}\\
      \multicolumn{2}{c}{P22: $56$ slices} & \multicolumn{2}{c}{P26: $38$ slices} & \multicolumn{2}{c}{P30: $98$ slices}\\
    \end{tabular}
    \caption{Test dataset with examples, identification number of the patients and number of slices related to the aneurysmatic tracts.}
    \label{test_dataset}
\end{figure}

From ROIs are created the binary masks for each slice of the sequences both for basal and with CM. \\Masks are produced with elliptical selection to include the entire vessel, and the pixels within the selected ellipse assumes as intensity value $1$ (white), while all the remaining ones takes $0$ (black). 
\\At this point, the dataset is ready for the processing in the portal ImageLab. The latter, takes as input the basal ROI and its corresponding CM one, together with their respective binary masks. These, are used by the algorithm to create the image $C$ (see Fig. \ref{schema_imageLab}), and consequently to detect the vessel. As depicted in Fig. \ref{interfaccia}, in output the extracted image is provided, along with a colored map where green highlights pixels correctly classified by the algorithm, red indicates misclassified ones, and white indicates the areas the algorithm should have identified but did not. Adjacent to this, similarity indices described before are presented, computed by considering the CM image as the target, which is provided as input for this purpose. We have processed in the portal each slice of our dataset.\\
As shown in Figure \ref{test_dataset}, for our tests we selected six patients with various types of aneurysms: with partially patent lumen (P19), floating thrombi (P14), with calcium plaques, nearly entire patent lumen (P22, P20, P26), and a case of aortic dissection (P30) which is a critical case of AAA that occurs when the wall of the vessel collapses, generating a "false lumen".

\subsection{Results of the DA}
We now report the results obtained from processing the six sequences on the ImageLab portal. 
In Table \ref{medie}, we can see the average values of the indices obtained from the elaborations, for every patient.\\ We can appreciate that the values of the \textit{DCI} index are rather high, indicating that on average, in the total dataset, $75\%$ of the white pixels are correctly classified. Furthermore, we can observe a tendency of the algorithm to underestimate the prediction of the patent lumen, with a total mean value of the $B_{pn}$ index of $-0.50$. From a mathematical standpoint, the $B_{pn}$ is optimal when it approaches zero in absolute value. If it takes positive values, it indicates an overestimation, whereas if it is negative, it indicates an underestimation, which is preferable from a medical perspective, because more conservative. \\Regarding the standard deviations, shown in Table \ref{std}, we notice that they are very small for each index, indicating that the algorithm works regularly, except for the $B_{pn}$ index, which evidently has a more significant variability, indicating that the algorithm varies greatly from image to image in overestimating or underestimating the vessel's lumen.
\begin{table}[H]
    \centering
    \begin{minipage}[t]{0.45\textwidth}
        \centering
        \begin{tabular}{|c|c|c|c|c|}
        \multicolumn{5}{c}{\textbf{\textit{Average values}}}\\
            \hline
            \textit{P.}  &  \textit{DCI} & \textit{TI} & \textit{Em} &  $B_{pn}$ \\
            \hline
            P14 &  0.76 & 0.62 & 0.38 & -0.49 \\
            \hline
            P19 & 0.71 & 0.56 & 0.45 & -0.18 \\
            \hline
            P20 & 0.73 & 0.57 & 0.43 & -0.69 \\
            \hline
            P22 & 0.74 & 0.59 & 0.41 & -0.74 \\
            \hline
            P26 & 0.77 & 0.63 & 0.37 & -0.60 \\
            \hline
            P30 & 0.79 & 0.65 & 0.35 & -0.53 \\
            \hline \hline
            Tot. & 0.75 & 0.60 & 0.39 & -0.50 \\ \hline
        \end{tabular}
        \caption{\small{Average indices for patients processed with ImageLab.}}
        \label{medie}
    \end{minipage}%
    \hspace{0.05\textwidth}
    \begin{minipage}[t]{0.45\textwidth}
        \centering
        \begin{tabular}{|c|c|c|c|c|}
        \multicolumn{5}{c}{\textbf{\textit{Standard deviations}}}\\
            \hline
            \textit{P.}  &  \textit{DCI} & \textit{TI} & \textit{Em} &  $B_{pn}$ \\
            \hline
            P14 &  0.06 & 0.08 & 0.08 & 0.49 \\
            \hline
            P19 & 0.06& 0.07 &0.07 &0.38 \\
            \hline
            P20 & 0.02& 0.02 & 0.02 & 0.08 \\
            \hline
            P22 & 0.08& 0.08 & 0.08 & 0.49 \\
            \hline
            P26 & 0.07& 0.08 & 0.08 & 0.28 \\
            \hline
            P30 & 0.03& 0.04 & 0.04 & 0.12 \\
            \hline\hline 
            Tot. & 0.05 & 0.06 & 0.06 & 0.30\\
            \hline
        \end{tabular}
        \caption{\small{Standard deviations for patients processed with ImageLab.}}
        \label{std}
    \end{minipage}
\end{table}
In Table \ref{dte_best} we show the best \textit{DCI}, \textit{TI} and \textit{Em} indices; in Table \ref{bpn_best} we report the slices in which the $B_{pn}$ performs better, i.e. is closer to $0$.
\begin{table}[H]
    \centering
    \begin{tabular}{ c c c c c c c}
    \hline
        \textit{Patient} & \textit{Basal} & \textit{CM} & \textit{Processed} & \textit{DCI} & \textit{TI} & \textit{Em} \\ \hline
        P14 & \includegraphics[width=1.0cm]{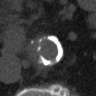} & \includegraphics[width=1.0cm]{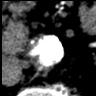} & \includegraphics[width=1.0cm]{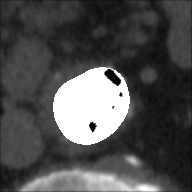}& 0.894 & 0.809 & 0.191 \\  
        \hline \vspace{0.1cm} 
         P19 & \includegraphics[width=1.0cm]{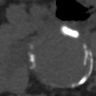} & \includegraphics[width=1.0cm]{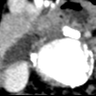} & \includegraphics[width=1.0cm]{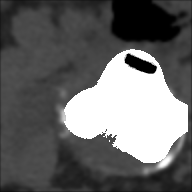}& 0.829 & 0.708 & 0.292 \\ 
        \hline \vspace{0.1cm}
         P20 & \includegraphics[width=1.0cm]{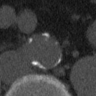} & \includegraphics[width=1.0cm]{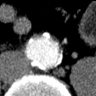} & \includegraphics[width=1.0cm]{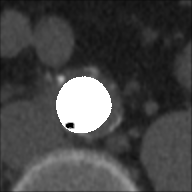}& 0.778 & 0.637 & 0.363 \\ 
        \hline \vspace{0.1cm}
        P22 & \includegraphics[width=1.0cm]{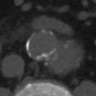} & \includegraphics[width=1.0cm]{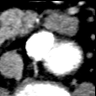} & \includegraphics[width=1.0cm]{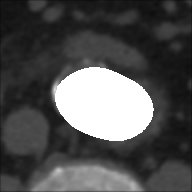}& 0.838 & 0.721 & 0.279 \\ 
        \hline \vspace{0.1cm}
        P26 & \includegraphics[width=1.0cm]{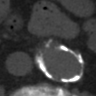} & \includegraphics[width=1.0cm]{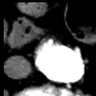} & \includegraphics[width=1.0cm]{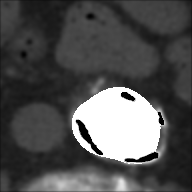}& 0.877 & 0.781 & 0.219 \\ 
        \hline \vspace{0.1cm}
        P30 & \includegraphics[width=1.0cm]{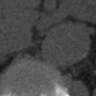} & \includegraphics[width=1.0cm]{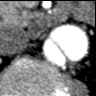} & \includegraphics[width=1.0cm]{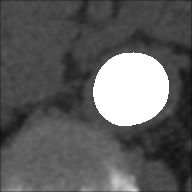}& 0.846 & 0.733 & 0.267 \\ 
        \end{tabular}
    \caption{\small{Best result based on \textit{DCI}, \textit{TI} and \textit{Em}. In the seven columns are represented respectively: the identification number of the patient, the basal image, the CM one, the processed image in which we highlighted in black the calcium plaques}, the \textit{DCI}, \textit{TI} and \textit{Em} indices.}
    \label{dte_best}
\end{table}
\vspace{-0.6cm}
\begin{table}[H]
\centering
    \begin{tabular}{ c c c c c c }
    \hline
    \textit{Patient} & \textit{Basal} & \textit{CM} & \textit{Processed} & \textit{Superposed} & $B_{pn}$ \\ \hline
    P14& \includegraphics[width=1.0cm]{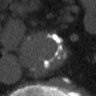} & \includegraphics[width=1.0cm]{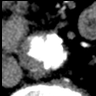} & \includegraphics[width=1.0cm]{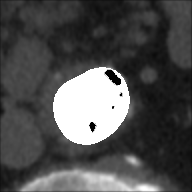}& \includegraphics[width=1.0cm]{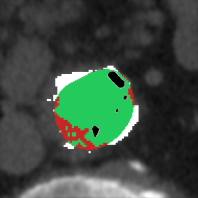} & -0.013 \\ 
    \hline \vspace{0.1cm} 
         P19 & \includegraphics[width=1.0cm]{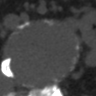} & \includegraphics[width=1.0cm]{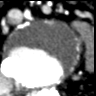} & \includegraphics[width=1.0cm]{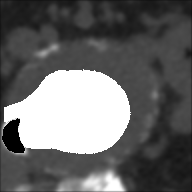}& \includegraphics[width=1.0cm]{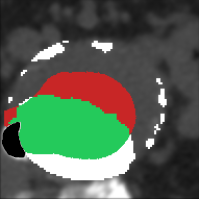} & -0.003 \\ 
    \hline \vspace{0.1cm}
        P20 & \includegraphics[width=1.0cm]{P20/BASALE_077.png} & \includegraphics[width=1.0cm]{P20/MDC_077.png} & \includegraphics[width=1.0cm]{P20/prev_077.png}& \includegraphics[width=1.0cm]{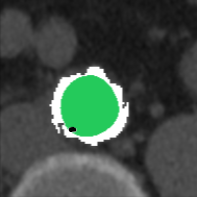} & -0.569 \\ 
        \hline \vspace{0.1cm}
        P22 & \includegraphics[width=1.0cm]{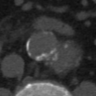} & \includegraphics[width=1.0cm]{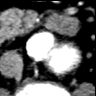} & \includegraphics[width=1.0cm]{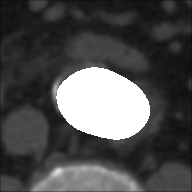}& \includegraphics[width=1.0cm]{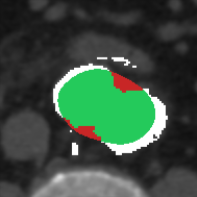} & -0.191 \\ 
        \hline \vspace{0.1cm}
        P26 & \includegraphics[width=1.0cm]{P26/BASALE_018.png} & \includegraphics[width=1.0cm]{P26/REGISTERED_18.png} & \includegraphics[width=1.0cm]{P26/018.png}& \includegraphics[width=1.0cm]{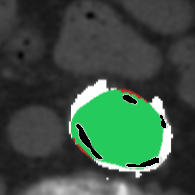} &-0.235 \\ 
        \hline \vspace{0.1cm}
        P30 & \includegraphics[width=1.0cm]{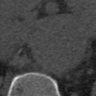} & \includegraphics[width=1.0cm]{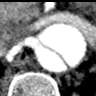} & \includegraphics[width=1.0cm]{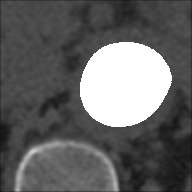}& \includegraphics[width=1.0cm]{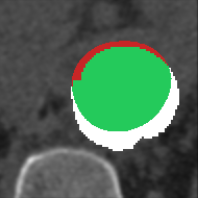}&-0.222 \\ 
    \end{tabular}
    \caption{\small{Best results based on $B_{pn}$. In the six columns we can see respectively: the identification number of the patient, the basal image, the corresponding CM one, the processed image where in black we highlighted the calcium plaques}, the coloured image showing the performance of the processing (superposed), see \cite{travaglini}, and the $B_{pn}$ index. In green are represented the pixels correctly classified by the algorithm, while in red, those classified incorrectly. Finally, in white are represented the pixels highlighted in the CM image but not identified by the algorithm.}
    \label{bpn_best}
    \end{table}

\begin{figure}[H]
        \hspace{-1cm}
        \begin{tabular}{c c c c c c}
         \textit{\textbf{Basal image}} & \textit{\textbf{CM image}} & \textit{\textbf{Version}} & \textbf{\textit{Extracted}} & \textit{\textbf{Colored map}} & \textit{\textbf{Indices}} \\ \\
         \multirow{8}{*}{\includegraphics[width=1.5cm]{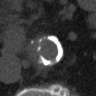}} & \multirow{8}{*}{\includegraphics[width=1.5cm]{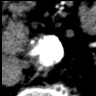}}& \multirow{4}{*}{Old} & \multirow{4}{*}{\includegraphics[width=1.5cm]{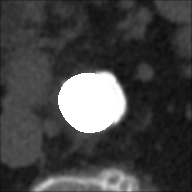}} & \multirow{4}{*}{\includegraphics[width=1.5cm]{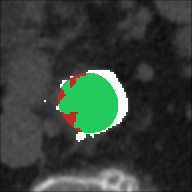}} & \textit{DCI}: 0.824 \\ &&&&& \textit{TI}: 0.700\\ &&&&& \textit{Em}: 0.300 \\ &&&&& $B_{pn}$: -0.192 \\ \\
           && \multirow{4}{*}{New} & \multirow{4}{*}{\includegraphics[width=1.5cm]{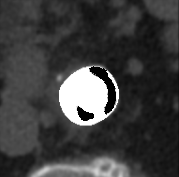}} & \multirow{4}{*}{\includegraphics[width=1.5cm]{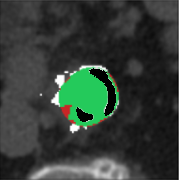}} & \textit{DCI}: 0.894 \\ 
           &&&&& \textit{TI}: 0.809\\ &&&&& \textit{Em}: 0.191 \\ &&&&& $B_{pn}$: -0.075\\
          \end{tabular}
        \caption{\small{Comparison in the processing procedure between the results obtained from the previous version of the algorithm (Old), see \cite{travaglini}, and the updated version used in this study (New). In the first column we present the basal image, and in the second, we show the target image. The third column specifies the version of the algorithm used. The next two columns display the extraction of the patent lumen and the colored map to evaluate the performance. Next to these, we include the similarity indices related to the processing.}}
        \label{P19_placche}
    \end{figure}

\begin{figure}[H]
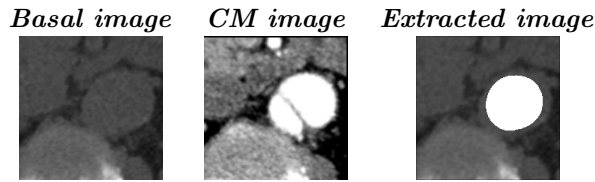

    \centering
    \begin{tabular}{c c c}
    \textit{\textbf{Basal image}} & \textbf{\textit{CM image}} & \textbf{\textit{Extracted image}}\\
       \includegraphics[width=1.9cm]{P30/BASALE_059.png}  & \includegraphics[width=1.9cm]{P30/REGISTERED_59.png} &      \includegraphics[width=1.9cm]{P30/059.png} \\
    \end{tabular}
    \caption{\small{Example of processed slice belonging to the patient affected from aortic dissection.}}
    \label{dissecazione}
\end{figure}

\begin{figure}[H]
    \begin{tabular}{c c | c c | c c }
    \multicolumn{2}{c|}{CM image}& \multicolumn{2}{c|}{CM image}& \multicolumn{2}{c}{CM image}\\
    \multicolumn{2}{c|}{\includegraphics[width=1.5cm]{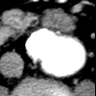}} & \multicolumn{2}{c|}{\includegraphics[width=1.5cm]{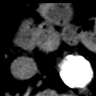}} & \multicolumn{2}{c}{\includegraphics[width=1.5cm]{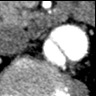}}\\ \multicolumn{2}{c|}{}&\multicolumn{2}{c|}{}&\multicolumn{2}{c}{}\\
   
    \multicolumn{2}{c|}{Extracted} & \multicolumn{2}{c|}{Extracted} & \multicolumn{2}{c}{Extracted} \\ 
   
       SK  & \hspace{-0.5cm} no SK & SK & \hspace{-0.5cm} no SK & SK & \hspace{-0.5cm} no SK \\
       \includegraphics[width=1.5cm]{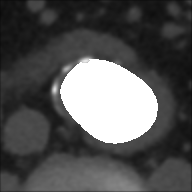}  & \includegraphics[width=1.5cm]{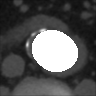} &  \includegraphics[width=1.5cm]{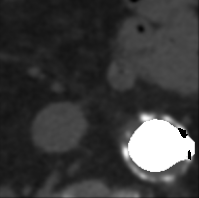} & \includegraphics[width=1.5cm]{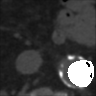} & \includegraphics[width=1.5cm]{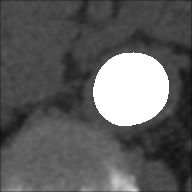} & \includegraphics[width=1.5cm]{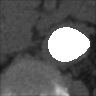}\\
\includegraphics[width=1.5cm]{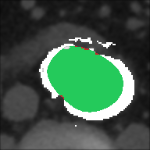} & \includegraphics[width=1.5cm]{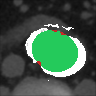} &  \includegraphics[width=1.5cm]{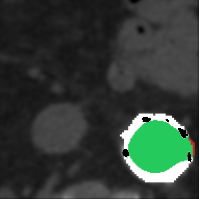} & \includegraphics[width=1.5cm]{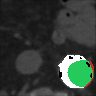} & \includegraphics[width=1.5cm]{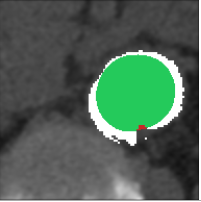} & \includegraphics[width=1.5cm]{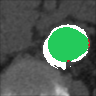}\\
       DCI: 0.814 & \hspace{-0.3cm} DCI: 0.807 & DCI: 0.788 & \hspace{-0.3cm} DCI:0.730 & DCI: 0.846 & \hspace{-0.3cm} DCI: 0.833\\
       TI: 0.686 & \hspace{-0.3cm} TI: 0.676 & TI: 0.651 & \hspace{-0.3cm} TI: 0.574 & TI: 0.733 & \hspace{-0.3cm} TI: 0.714\\
       Em: 0.314 & \hspace{-0.3cm} Em:0.324 & Em: 0.349 & \hspace{-0.3cm} Em: 0.426 & Em: 0.267 & \hspace{-0.3cm} Em: 0.286 \\
       $B_{pn}$: -0.440 & \hspace{-0.3cm} $B_{pn}$: -0.436 & $B_{pn}$: -0.501 & \hspace{-0.3cm}$B_{pn}$:-0.573 & $B_{pn}$:-0.349 & \hspace{-0.3cm} $B_{pn}$:-0.370\\
    \end{tabular}
    \caption{\small{Examples of results from the segmentation algorithm with and without reconstruction using SK operators. In the first row is depicted the CM image, i.e. the target. Below, we report the extracted images with (SK) and without SK (no SK). We also include the colored map and the corresponding similarity indices.}}
    \label{sk_vs_nosk}
\end{figure}

In this version of the algorithm (updated compared to the one described in \cite{travaglini}), the calcium plaques are excluded from the patent lumen of the vessel and therefore visible in the processed image (in black), as we can observe also from Tables \ref{dte_best} and \ref{bpn_best}. The advantage of removing these structures during the extraction process leads to a more accurate and faithful evaluation of the indices. This observation is illustrated in Figure \ref{P19_placche}, where we present a comparison of results for a single slice obtained using the previous algorithm version, which does not remove calcium plaques from both the basal and the CM images, and the new algorithm version. The performance improvement is evident, particularly in the \textit{DCI} index, where a difference is observed: $82\%$ of white pixels are accurately classified with the old version, compared to $89\%$ with the new version.  
Finally we highlight how, in the new version of the algorithm, the removal of the calcium plaques from both the basal and the CM images produces a different extraction of the patent lumen of the vessel, which in addition to being compatible with the algorithm's steps, also turns out to be more accurate.
\vspace{-0.3cm}
Another aspect is how the algorithm handles the critical case of dissection. As we can observe in Figure \ref{dissecazione}, it does not recognize the true lumen from the false lumen but treats it as if it was a typical aneurysm case.
Before moving on to the description of the second approach, it is interesting to highlight that in the previously described algorithm, the reconstruction with SK operators ensures overall better performances compared to the analogous algorithm executed without these operators. The differences can be observed in Fig. \ref{sk_vs_nosk}, where we can also appreciate from the obtained indices values that the algorithm with sampling Kantorovich operators works better in the extraction of the patent lumen of the vessel. In the first two columns, we have cases of typical aneurysms; in the third column we encounter a case of dissection. Although not recognized as such, as mentioned earlier, the algorithm generates good indices values by considering it as a case of a normal aneurysm. From the indices, we can see that the algorithm with SK reconstruction performs better than the same process without these operators.

\section{Artificial Intelligence approach (AI)}
In this section we present the second approach adopted, which relies on artificial intelligence. The objective is analogous to that of the DA: to obtain, starting from the basal image, the prediction of the patent lumen of the vessel. We trained a neural network that predicts the image with the contrast medium effect, taking the basal image as input.
\subsection{Description of the U-Net model}
\begin{figure}[H]
    \centering
\includegraphics[width=13cm, keepaspectratio]{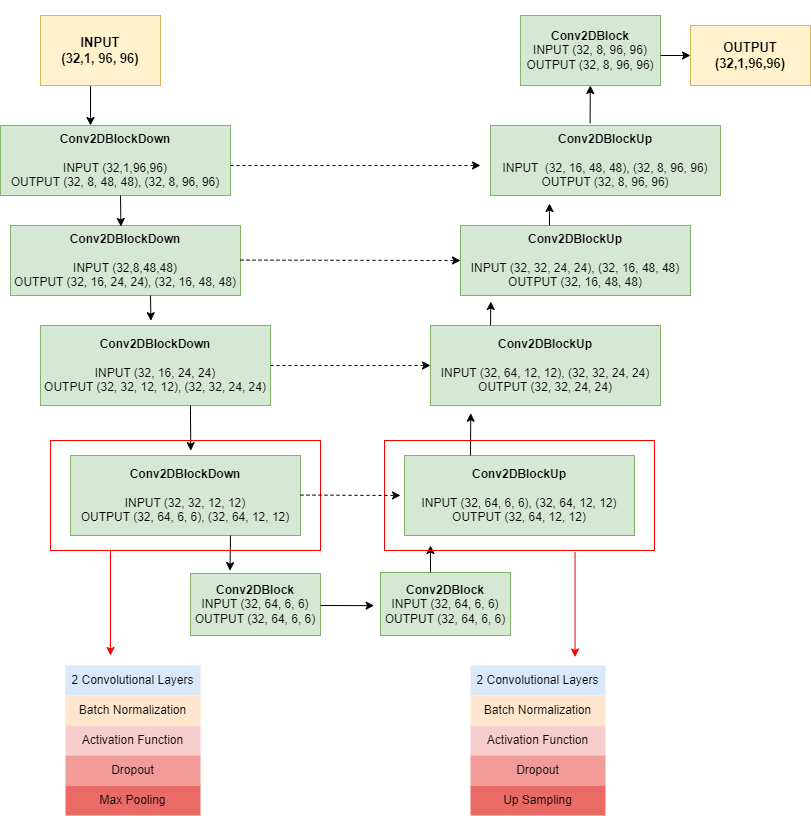}
    \caption{\small{Architecture of our U-Net.  }}
    \label{u_net}
\end{figure}

The neural network model that we have chosen is the one of U-Net \cite{olaf}, due to its particular suitability for image segmentation tasks. 
\\In Figure \ref{u_net}, the architecture of our U-Net is depicted, highlighting the composition of the encoding path (left branch) and decoding path (right branch). \\ Describing Figure \ref{u_net}, the input and output tensors has four coordinates which represents, respectively: the batch dimension (of 32 images), the channels, i.e. the number of filters that are applied to allow the network to extract features during the learning process, the size $(m,n)$ of the image that in our case is of $96\times96$ pixels. The encoding path consists of four convolutional blocks called "Conv2DBlockDown", each of which contains: two convolutional layers, batch normalization operations to reduce variability, engagement of the activation function, random deactivation of some neurons through Dropout operation, and MaxPooling operations that halve the dimensions of the image by synthesizing its informational content (for details, see e.g. \cite{goodfellow}). The decoding path is symmetric with four convolutional blocks "Conv2DBlockUp", with the only difference being in the Up Sampling operation in each block, which reconstructs the image in terms of both dimensions and informational content. In the outputs of the blocks composing the encoding path, we have two tensors: the first is the real output of the block, on which convolution has been applied, while the second one is the tensor used for the corresponding skip connection, which contribute to better accuracy of the results. In the opposite way for the decoding path, where we have two tensors in input and one tensor in output.
In what follows we resume the learning process of the U-Net.
\begin{quote}
\textbf{\underline{TRAINING:}}
    $200$ epochs, for batches of $32$ images.\\ \\
    Objective: Minimizing the Loss function \begin{equation}
\label{loss}
    L_{SSIM}(Y, Y_{p}):=1-SSIM(Y, Y_{p}),
\end{equation}
\hspace{1.7cm}in which we used the $SSIM$ defined in Section $3.1$.\\ 

\hspace{0.5cm}\underline{Training phase:}
\begin{quote}
    \begin{itemize}
        \item Selection of the input data $X$ and the target $Y$ from the batch;
        \item Prediction of the output $Y_{p}$;
        \item Computation of the Loss function $L_{SSIM}(Y, Y_{p})$;
        \item Optimization of the weights adopting the \textit{AdamW} optimization algorithm, see e.g. \cite{adamw};
        \item Regulation of the learning rate, experimentally determined after several tests, with an initial value of $1 \cdot 10^{-6}$ and an upper bound of $1 \cdot 10^{-3}$, to achieve optimal results. For this procedure we used the \textit{OneCycleLR} optimizer, see e.g. \cite{onecycle}. 
    \end{itemize}
    \begin{quote}These steps are iterated until every batch has been processed within the epoch. Once this is achieved, the validation phase begins.
    \end{quote}
    \end{quote}
   \hspace{0.5cm} \underline{Validation phase:}
   \begin{quote}
    \begin{itemize}
        \item Extraction of the input from the batch within the validation dataset;
        \item Prediction step with the generation of $Y_{p}$;
        \item Computation of the Loss function using the weights obtained during the training phase.
    \end{itemize}
    \end{quote}
    \textbf{\underline{TESTING:}} Production of the output $Y_{p}$.
\end{quote}

\subsection{Description of the dataset}
Our dataset consists of $25$ sequences of CT images of patient affected by abdominal aorta aneurysm, obtained in the same way as the DA.  
Each stage of the learning process has a dedicated dataset. Therefore, we have one dataset for the training phase ($15$ patients, $992$ slices), one for validation ($4$ patients, $277$ slices), and one for testing ($6$ patients, $362$ slices), the latter being the same dataset used to test the segmentation algorithm of the deterministic approach. We have organized the datasets into three different folders.
Our network is trained for batches of 32 images taken from the appropriate dataset folder. The images in the datasets are organized into folders divided by patient. Within each patient's folder, there are two subfolders: one containing the sequences of images taken in basal mode, and the other containing the sequences of images taken with contrast medium.\\Images are properly registered and cropped to obtain the ROIs of dimensions $96\times 96$, following the same procedure of the DA. The only exception is for a patient for which was necessary a crop of bigger dimensions ($192\times 192$ pixels), in order to include completely the aneurysm through all the sequence. At this point images undergo to normalization of the intensities, scaling to $96\times96$ pixel dimensions (when necessary), data augmentation through rotations and reflections and finally conversion of data into tensors. After these pre-processing steps, dataset is ready for the net. 

\subsection{Results of the AI approach}
\begin{figure}[H]
    \centering
    \includegraphics[width=12cm, keepaspectratio]{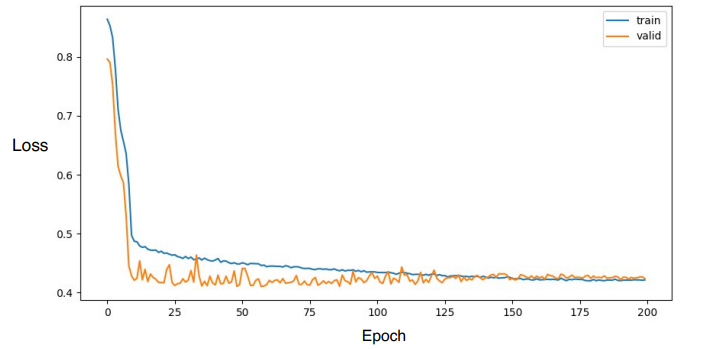}
    \caption{\small{Loss function respectively of the training (in blue), and the validation phase (in orange).}}
    \label{grafico loss}
\end{figure}

For the U-Net training we used a workstation with processor Intel i9-13900K, $64$GB of ram, GPU NVIDIA RTX A2000 with $12$GB of VRAM. \\The training execution time is about $7$ minutes and $33$ seconds.\\
The first result returned by the network is the graph of the performances of the Loss functions, see Figure \ref{grafico loss}.
Interpreting the graph, we can notice how in the initial epochs both losses decrease rapidly, indicating that the model is suitable for the problem. It can be observed that the validation loss is slightly lower than the training loss, possibly indicating that the validation dataset is relatively easy for the network, which is typical of the early epochs. From approximately the $100$-th epoch onwards, the two functions converge to a common minimum until the end of the epochs. This behavior allows us to conclude that the training was successful, meaning that the network generalizes well and there are no problematic phenomena such as excessive model complexity (overfitting) or excessive simplicity (underfitting).

\begin{figure}[H]
\hspace{-1.5cm}
\begin{tabular}{c c c c || c c c c }
\textbf{\textit{Patient}} & \textbf{\textit{Basal}} & \textbf{\textit{CM}} & \textbf{\textit{Output}} & \textbf{\textit{Patient}} & \textbf{\textit{Basal}} & \textbf{\textit{CM}} & \textbf{\textit{Output}}\\
   P14 & \includegraphics[width=1.5cm]{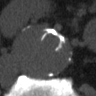}  & \includegraphics[width=1.5cm]{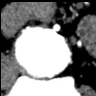} & \includegraphics[width=1.5cm]{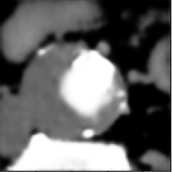} & 
P22 & \includegraphics[width=1.5cm]{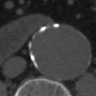} & \includegraphics[width=1.5cm]{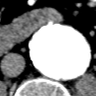} & \includegraphics[width=1.5cm]{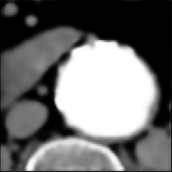} \\
    P19 & \includegraphics[width=1.5cm]{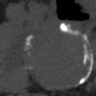} & \includegraphics[width=1.5cm]{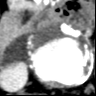} & \includegraphics[width=1.5cm]{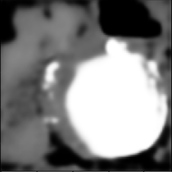} &
   P26 & \includegraphics[width=1.5cm]{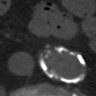} & \includegraphics[width=1.5cm]{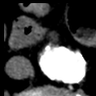} & \includegraphics[width=1.5cm]{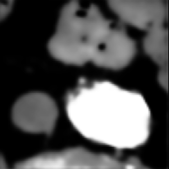}\\
   P20 & \includegraphics[width=1.5cm]{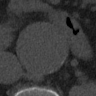} & \includegraphics[width=1.5cm]{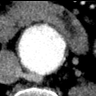} & \includegraphics[width=1.5cm]{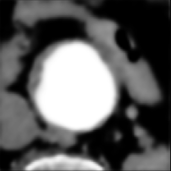} &
   P30 & \includegraphics[width=1.5cm]{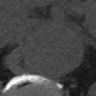} & \includegraphics[width=1.5cm]{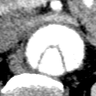} & \includegraphics[width=1.5cm]{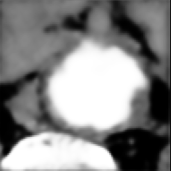}\\
\end{tabular}
\caption{Examples of output provided by the U-Net, for each patient.}
    \label{esempi_output_rete}
\end{figure}

In Figure \ref{esempi_output_rete} we show samples of prediction, i.e. the output, provided by the net, for each patient.

We can notice how the network's output is accurate on pattern recognition, although it is still very blurry. This issue may be correlated with the modest dimension of the dataset and the complexity of the problem.\\ \\
After obtaining the predictions $Y_{p}$ from the network,binarization is performed with an experimentally determined optimal threshold value of $0.8$, and masks representing the calcium plaques are calculated, as in the deterministic approach. These steps allow for the computation of the \textit{DCI, TI, Em}, and $B_{pn}$ indices and the subsequent comparisons. 

Now, in the following Table \ref{average_net} we can see the mean values through all the test dataset for the indices \textit{DCI, TI, Em, Bpn}, and also the \textit{SSIM} that the net computed during the test phase. We note that, the latter index is evaluated between the target image and the output of the net.
\begin{table}[H]
    \centering
    \begin{tabular}{|c|c|c|c|c|}
    \hline
         \textbf{\textit{DCI}} & \textbf{\textit{TI}} & \textit{\textbf{Em}} & \textit{\textbf{Bpn}} & \textbf{\textit{SSIM}}\\ 
    \hline
         0.81 & 0.70 & 0.30 & -0.05 & 0.46\\
    \hline
    \end{tabular}
    \caption{\small{Average values of the indices computed by the net.}}
    \label{average_net}
\end{table}
From the \textit{DCI} analysis, it can be deduced that $81\%$ of the white pixels are well classified. Moreover, there is a slight underestimation in forecasting the patent lumen in comparison to the target images, as evidenced by the \textit{Bpn} index. The \textit{SSIM} score is relatively low, and it is consistent with the network's output. This is primarily due to the fact that in these images, pixel classification often deviates from the targets, which constitutes a significant factor impacting the \textit{SSIM} value. Furthermore, the blurring present in the network outputs significantly affect the value of this index.

\section{Comparison}
\begin{figure}[H]
    \centering
    \begin{tabular}{c c c c c}
       Patient & Basal & CM & DA & AI \\ \hline 
         P14& \includegraphics[width=1.5cm]{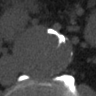}& \includegraphics[width=1.5cm]{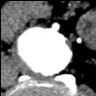} & \includegraphics[width=1.5cm]{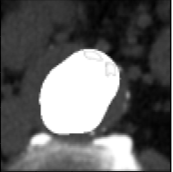} & \includegraphics[width=1.5cm]{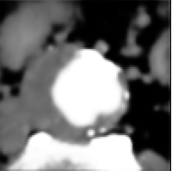} \\ 
         & & &  \scriptsize{DCI: $0.804$} & \scriptsize{DCI: $0.662$}\\
         \hline
         P19 & \includegraphics[width=1.5cm]{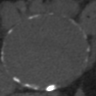} & \includegraphics[width=1.5cm]{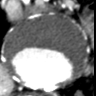} & \includegraphics[width=1.5cm]{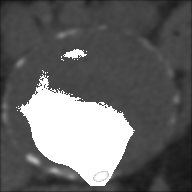} & \includegraphics[width=1.5cm]{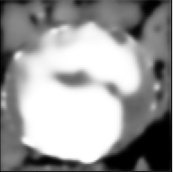} \\ 
         & & & \scriptsize{DCI: $0.818$} & \scriptsize{DCI: $0.642$}\\
         \hline
         P20 & \includegraphics[width=1.5cm]{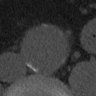} & \includegraphics[width=1.5cm]{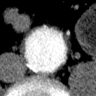} & \includegraphics[width=1.5cm]{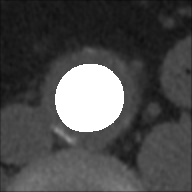} & \includegraphics[width=1.5cm]{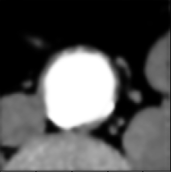} \\ 
         & & & \scriptsize{DCI: $0.817$} & \scriptsize{DCI: $0.947$}\\
         \hline
          P22 & \includegraphics[width=1.5cm]{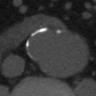} & \includegraphics[width=1.5cm]{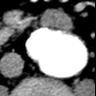} & \includegraphics[width=1.5cm]{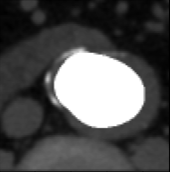} & \includegraphics[width=1.5cm]{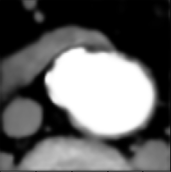} \\ 
        & & & \scriptsize{DCI: $0.824$} & \scriptsize{DCI: $0.929$}\\
        \hline
        P26 & \includegraphics[width=1.5cm]{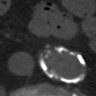} & \includegraphics[width=1.5cm]{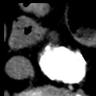} & \includegraphics[width=1.5cm]{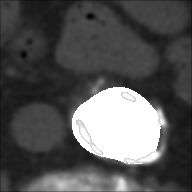} & \includegraphics[width=1.5cm]{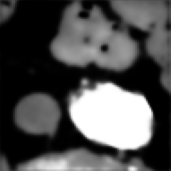} \\ 
        & & & \scriptsize{DCI: $0.918$} & \scriptsize{DCI: $0.923$}\\
        \hline 
        P30 & \includegraphics[width=1.5cm]{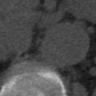} & \includegraphics[width=1.5cm]{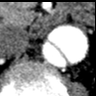} & \includegraphics[width=1.5cm]{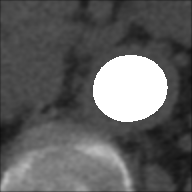} & \includegraphics[width=1.5cm]{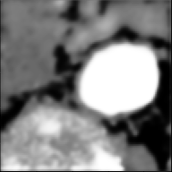}\\ 
        & & & \scriptsize{DCI: $0.845$} & \scriptsize{DCI: $0.837$}\\
        \hline
    \end{tabular}
    \caption{\small{Comparison of outputs with their respective \textit{DCI} indices.}}
    \label{confronto_output}
\end{figure}

\begin{figure}[H]
    \includegraphics[width=6.5cm, keepaspectratio]{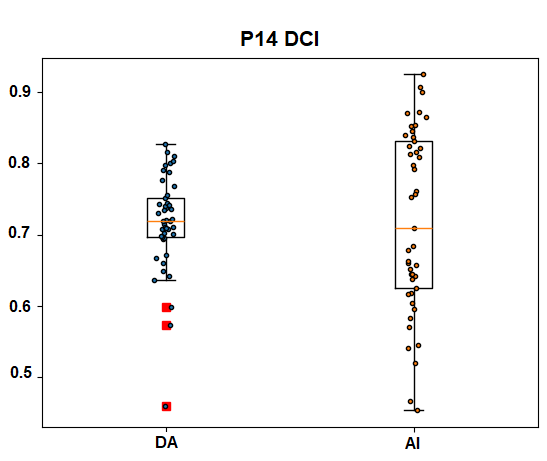} 
    \includegraphics[width=6.5cm, keepaspectratio]{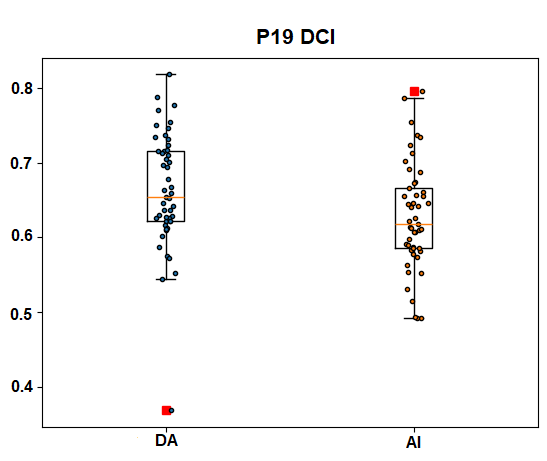}
    \includegraphics[width=6.5cm, keepaspectratio]{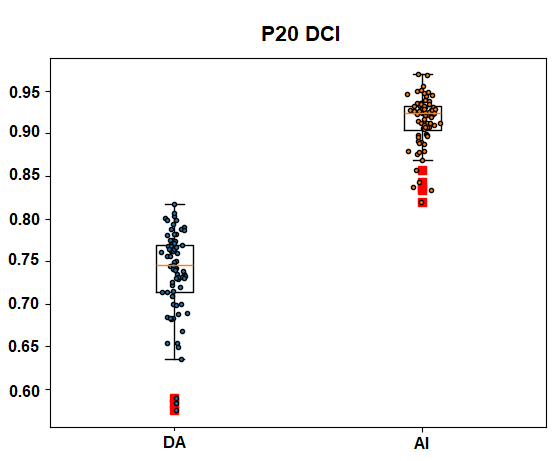} \vspace{0.3mm}
    \includegraphics[width=6.5cm, keepaspectratio]{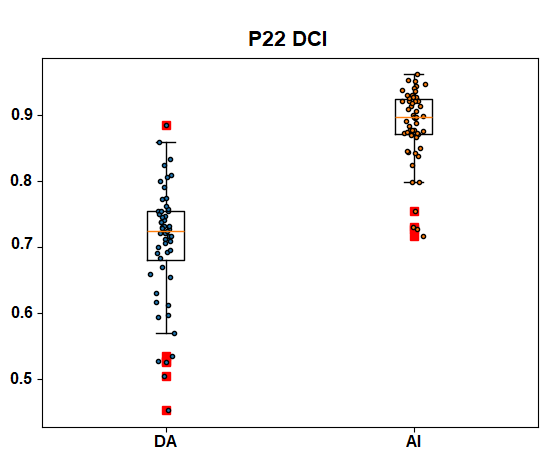}
    \includegraphics[width=6.5cm, keepaspectratio]{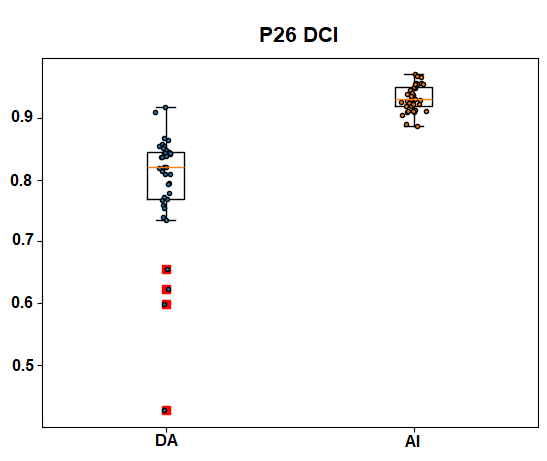}
    \includegraphics[width=6.5cm, keepaspectratio]{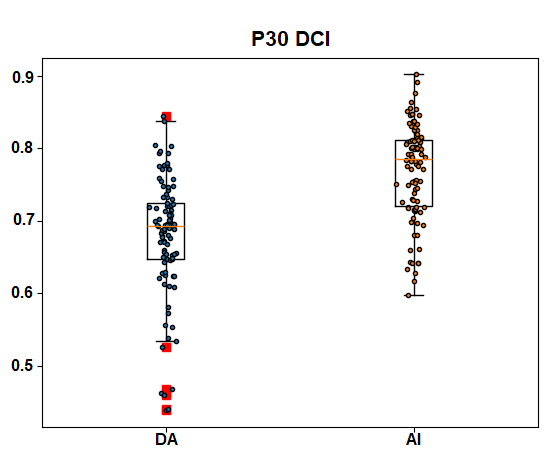}
    \caption{\small{Comparisons of 
    \textit{DCI} index between DA and AI approach by means of boxplots. The dots represent the values of \textit{DCI} obtained for each slice, in blue for the DA and in orange for the AI approach. In general, the latter assumes higher values, with less variation in the distribution and fewer outliers (highlighted in red). Where variability is significant, we encounter patients with complex morphologies that are challenging to process. Conversely, uniform and compact \textit{DCI} distributions indicate patients easily processed by the models.}}
    \label{confronto_dci}
\end{figure}

In this section, we present the comparison of the results obtained by the two methods.
This involves computing the similarity indices defined previously in the discussion.
Consequently, we have identified significant statistics to determine which of the two approaches performs better.
In Table \ref{statistiche_confronto} we resume the main statistical values for the indices used for the evaluations in both the two methods.
\begin{table}[H]
    \centering
    \begin{tabular}{|c|c c|c c|c c|c c|}
    \cline{2-9}
       \multicolumn{1}{c|}{} & \multicolumn{2}{|c|}{\textbf{\textit{DCI}}} & \multicolumn{2}{|c|}{\textbf{\textit{TI}}} & \multicolumn{2}{|c|}{\textbf{\textit{Em}}} & \multicolumn{2}{|c|}{\textbf{\textit{Bpn}}} \\ 
        \cline{2-9}
        \multicolumn{1}{c|}{} & \textbf{DA} & \textbf{AI} & \textbf{DA}& \textbf{AI} & \textbf{DA} & \textbf{AI} & \textbf{DA} & \textbf{AI} \\ [1ex] \hline 
        \textbf{mean}  & 0.71 & 0.81 & 0.56 & 0.70 & 0.44 & 0.30 & -0.66 & -0.05 \\ [1ex] \hline 
        \textbf{std}  & 0.08 & 0.12 & 0.10 & 0.17 & 0.10 & 0.17 & 0.48 & 0.49 \\ [1ex] \hline 
        \textbf{min}  & 0.37 & 0.45 & 0.23 & 0.29 & 0.15 & 0.06 & -2.68 & -2.37 \\ [1ex] \hline 
        \textbf{25\%} & 0.66 & 0.72 & 0.49 & 0.56 & 0.38 & 0.15 & -0.83 & -0.21 \\ [1ex] \hline 
        \textbf{50\%} & 0.72 & 0.84 & 0.57 & 0.72 & 0.44 & 0.28 & -0.65 & -0.08 \\ [1ex] \hline 
        \textbf{75\%} & 0.77 & 0.92 & 0.62 & 0.85 & 0.50 & 0.44 & -0.43 & 0.06 \\ [1ex] \hline 
        \textbf{max}  & 0.92 & 0.97 & 0.85 & 0.94 & 0.77 & 0.70 & 0.76 & 1.17 \\ [1ex] \hline
    \end{tabular}
    \caption{\small{Comparison of statistics for the indices between the DA and the AI approach.}}
    \label{statistiche_confronto}
\end{table}
From Table \ref{statistiche_confronto}, it is evident that there exists a slight advantage for the neural network over the deterministic approach (DA) in terms of the \textit{DCI, TI}, and \textit{Em} indices. The neural network, on average, achieves an $81\%$ \textit{DCI} on the total test dataset, whereas the deterministic approach achieves $71\%$. However, it is noteworthy from the second percentile that the $50\%$ of the dataset exhibits \textit{DCI} values greater than $72\%$, indicating a promising performance. The maximum \textit{DCI} value of $92\%$ achieved by the deterministic approach is slightly lower than the maximum obtained with by the AI approach ($97\%$). \\
From the average values of the $B_{pn}$ index, it can be deduced that the deterministic approach tends to underestimate predictions compared to the AI approach. This index demonstrates significant variability compared to others. 
\\In Figure \ref{confronto_output} we provide some examples of outputs to give a visual comparison. We have also included the \textit{DCI} index, which is maybe the most intuitive for evaluating the performance of the two approaches.
We notice that, both approaches demonstrate effective performance. The output from the deterministic approach (DA) is clear, with regular contours of the patent lumen, in contrast to the AI approach output, which exhibits blurriness and more realistic contours. Occasional instances of bad performances in both approaches may be attributed to the complexity of the morphology of individual slices. Nonetheless, the overall quality of predictions is deemed satisfactory.
\\Finally, we provide in Figure \ref{confronto_dci} the boxplots showing the distribution of the DCI index, for each patient and for both the approaches.

\section{Conclusions}
We compared two approaches aimed at extracting the representation of the patent lumen of the abdominal aortic vessel in patients with AAA. Both methods yielded satisfactory results. The DA, using ImageLab (see e.g. \cite{travaglini}), has been improved with the addition of the removal of calcium plaques from both the basal and the target CM image, allowing more accurate computation of indices, further improving the performance of the DA. Conversely, despite the rather limited dataset, the neural network produced valid results, overall better than the deterministic model. Recalling the boxplots depicted in Figure \ref{confronto_dci}, it becomes evident that the AI approach operates with less variability and higher precision compared to the DA. However, it's noteworthy that the deterministic approach still yields commendable results. Thus, we observe a slight advantage of one approach over the other. \\One benefit of this AI procedure lies in its operator independence, once the ROI is selected from the CT image stack. In contrast, the DA requires additional preliminary steps, such as creating masks to which the portal ImageLab seems to be sensitive. On the other hand, a notable advantage of the deterministic approach is the ability to directly intervene in the algorithm, understanding all the steps required to produce the output, unlike the AI approach. The implementation and refinement of these methodologies could enhance medical imaging diagnostics and improve the prevention and treatment of certain pathologies, thereby positively impacting the quality of life for many patients.\\It would be interesting to further explore these approaches, and we are optimistic about potential enhancements that could further refine the performance of both the DA and the AI approach. Given the outcomes achieved, we believe this study holds promise in assisting with the identification of aortic vessel patency in cases of AAA.

\bibliographystyle{fancy}

\section*{Funding}
The fourth author has been supported within the project "Mathematical Methods and Models for Biomedical Applications" financied by National Recovery and Resilence Plan PNRR-III-C9-2022-I8.

\section*{Informed Consent Statement}
All subjects involved gave their informed consent for inclusion before they participated in the study. All the images used are completely anonymized and it is not possible to trace the patients from the data mentioned in the paper.

\section*{Conflict of interest/Competing interests}
The authors declare that they have no conflict of interest and competing interest.

\section*{Availability of data and material and Code availability}
Not applicable.

\end{document}